# Theoretical foundations of the integral indicator application in hyperparametric optimization

Roman S. Kulshin[1], Anatoly A. Sidorov[1]
[1]Tomsk State University of Control Systems and Radioelectronics
Tomsk, Russian Federation
roman.s.kulshin@tusur.ru

*Abstract* — The article discusses the concept of hyperparametric optimization of recommendation algorithms using an integral assessment that combines various performance indicators into a single consolidated criterion. This approach is opposed to traditional methods of setting up a single metric and allows you to achieve a balance between accuracy, ranking quality, variety of output and the resource intensity of algorithms. The theoretical significance of the research lies in the development of a universal multi-criteria optimization tool that is applicable not only in recommendation systems, but also in a wide range of machine learning and data analysis tasks.

## I. Introduction

Recommendation systems in the modern digital space are becoming crucial, providing personalized user access to information, products and services. Their quality and effectiveness directly determine the level of audience satisfaction and affect the business performance of companies implementing such technologies. However, building and maintaining recommendation algorithms involves a number of methodological challenges. One of them is the problem of setting hyperparameters, which determine the generalization ability of the model, its stability and performance.

For a long time, the hyperparameter optimization process has been based on maximizing individual metrics [1]. In the simplest case, the researcher chooses a target indicator, such as the accuracy or quality of the ranking, and strives to improve it. However, practical experience demonstrates the limitations of this approach. An improvement in one characteristic of a model often leads to a deterioration in others. An increase in accuracy is accompanied by a decrease in diversity, an increase in ranking quality can increase computational complexity, and a reduction in system response time can lead to a loss of issue relevance. Thus, there is a contradiction between individual quality indicators, which makes the task of optimizing hyperparameters multi-criteria.

Traditional methods of searching for optimal parameter values, such as grid search or random search, allow us to explore the solution space, but do not answer the fundamental question: by what criterion should the effectiveness of the model be evaluated if the criteria themselves contradict each other [2]. Against this background, the integral assessment, which combines indicators into a single composite index, becomes particularly important. It allows you to avoid excessive focus on one characteristic and take into account the cumulative behavior of the algorithm.

## II. Integral assessment as a universal criterion

An integral score is an aggregated indicator that includes several groups of metrics, each of which reflects a specific aspect of the model's operation. Unlike simple averaging, it is based on a specially developed methodology that includes data normalization, grouping by sub-indexes, and subsequent weighted aggregation.

The process of forming an integral indicator begins with the selection of metrics that reflect the accuracy of recommendations, the quality of their ranking, the variety and cost of computing resources. These characteristics together provide a holistic view of the recommendation system's functionality. However, the metric values are expressed in different units of measurement and are in disparate ranges. To eliminate this discrepancy, minimax normalization is applied, which translates all indicators into a single scale from zero to one. In the case of resource-intensive parameters, such as execution time and memory usage, inverse normalization is used, since an increase in their values indicates a deterioration in the quality of the system.

After normalization, the indicators are combined into sub-indexes: accuracy, ranking, diversity, and resources. Within each subindex, metrics are aggregated using weights calculated using the entropy method. This approach allows us to take into account the variability of each indicator and distribute the contribution to the integrated assessment objectively, without subjective intervention. As a result, a final index is formed, which acts as a universal criterion for comparing models and optimizing their parameters.

## III. Hyperparametric optimization methods in the context of integral estimation

The use of an integral indicator as an objective function opens up opportunities for using various hyperparameter optimization methods. Among them, grid search, random search, evolutionary strategies, and Bayesian approaches are traditionally distinguished. Each method has its own advantages and limitations.

Sorting through the grid ensures consistency, but it quickly becomes computationally expensive as the number of parameters increases. Random search allows you to cover a wider range of space, but it does not take into account the results of previous steps. Evolutionary strategies such as CMA-ES [3] are focused on finding the global optimum and take into account the relationships between the parameters, but require large resources. The most promising in the context of working with integral estimation are Bayesian methods, in particular the Tree-structured Parzen Estimator [4]. This method builds probabilistic models of "successful" and "unsuccessful" combinations of hyperparameters and allows you to focus on those areas of the search space that have the greatest potential to improve the final integral score.

Thus, the use of an integral indicator in combination with adaptive optimization methods creates the basis for effective model tuning that takes into account the complex quality of the algorithm.

## IV. Strategies for applying integrated assessment

The use of an integral indicator in the process of hyperparametric optimization allows you to implement several strategies. The first strategy is to maximize the integral index as such, which ensures a balanced improvement of all the characteristics of the model. This approach is especially useful in cases where it is necessary to create a universal system that works equally well in all key areas.

The second strategy is based on the introduction of a dominant subindex. In this case, the objective function is shifted towards a priority aspect, such as accuracy, but the rest of the indicators retain weight, which prevents them from falling critically. This technique turns out to be in demand in applied problems, where the emphasis is on one criterion, but ignoring others can lead to a decrease in the practical value of the model.

The third strategy involves focusing solely on one metric. It provides a significant increase in the target indicator, but is accompanied by a deterioration in other characteristics. Despite its high efficiency in specialized scenarios, this approach remains risky from the point of view of system stability. Comparative analysis shows that integrated assessment and subindex strategies ensure a more harmonious development of models, while isolated optimization often leads to an imbalance.

## V. Methodological challenges and development directions

Despite the obvious advantages of the integral approach, its practical application is fraught with certain difficulties. One of the key problems is the definition of weights. The entropy method allows you to automate the distribution of coefficients, but it can lead to underestimation of the significance of individual metrics. The solution may be a combined approach that takes into account both statistical characteristics and expert assessments.

Another challenge is the normalization problem. During the operation of the model, some metrics may exceed the set range, which creates distortions in the final estimate. A promising direction is the development of dynamic or adaptive normalization methods that can take into account the specifics of the data and adjust ranges depending on the behavior of the algorithm.

## VI. Conclusion

Integral evaluation as a hyperparametric optimization tool opens up new possibilities for building recommendation systems and other machine learning models. It allows you to abandon the narrowly focused focus on individual metrics and move on to a comprehensive quality assessment. This approach ensures balanced algorithm development, minimizes conflict between metrics, and makes the setup process more transparent and universal.

The presented concept is applicable not only in the field of recommendations, but also in the tasks of financial analysis, industrial optimization and engineering design, where conflicting criteria also need to be taken into account. Further development of the method is associated with the refinement of normalization and weighting procedures, as well as the study of its adaptation to various types of data and model architectures.

## Acknowledgment

This paper is designed as part of the state assignment of the Ministry of Science and Higher Education; project FEWM 2023-0013.